\documentclass[journal,twoside, web]{ieeecolor}
\usepackage{graphicx}
\usepackage{epsfig}
\usepackage{epstopdf}
\usepackage{generic}
\usepackage{cite}
\usepackage{amsmath}
\usepackage{multirow}
\usepackage{indentfirst}
\usepackage{amsmath,amssymb,amsfonts}
\usepackage{algorithmic}
\usepackage{multicol}
\usepackage{booktabs}
\usepackage{algorithm,algorithmic}
\usepackage{hyperref}
\usepackage{float}
\hypersetup{hidelinks=true}
\usepackage{textcomp}
\def\BibTeX{{\rm B\kern-.05em{\sc i\kern-.025em b}\kern-.08em
    T\kern-.1667em\lower.7ex\hbox{E}\kern-.125emX}}
\begin{document}
\title{Frequency Feature Fusion Graph Network For Depression Diagnosis Via fNIRS}
\author{Chengkai Yang, Xingping Dong, and Xiaofen Zong, \IEEEmembership{Member, IEEE}
}

\maketitle

\begin{abstract}
Data-driven approaches for depression diagnosis have emerged as a significant research focus in neuromedicine, driven by the development of relevant datasets. Recently, graph neural network (GNN)-based models have gained widespread adoption due to their ability to capture brain channel functional connectivity from both spatial and temporal perspectives. However, their effectiveness is hindered by the absence of a robust temporal biomarker.
In this paper, we introduce a novel and effective biomarker for depression diagnosis by leveraging the discrete Fourier transform (DFT) and propose a customized graph network architecture based on Temporal Graph Convolutional Network (TGCN). Our model was trained on a dataset comprising 1,086 subjects, which is over 10 times larger than previous datasets in the field of depression diagnosis. Furthermore, to align with medical requirements, we performed propensity score matching (PSM) to create a refined subset, referred to as the PSM dataset.
Experimental results demonstrate that incorporating our newly designed biomarker enhances the representation of temporal characteristics in brain channels, leading to improved F1 scores in both the real-world dataset and the PSM dataset. This advancement has the potential to contribute to the development of more effective depression diagnostic tools. In addition, we used SHapley Additive exPlaination (SHAP) to validate the interpretability of our model, ensuring its practical applicability in medical settings.
\end{abstract}


\section{Introduction}
        Depression is a common but serious mental health disorder that is spreading globally. More than 350 million people worldwide suffer from depression, which has become the fourth most common disease in the world. For a long time, the diagnosis of depression relied on experienced specialists.\par
        Recently, with the advancement of smart healthcare, brain imaging techniques such as Functional Magnetic Resonance Imaging (fMRI) \cite{groenewold2013emotional}, Electroencephalogram (EEG) \cite{zhang2020brain}, and Functional Near-Infrared Spectroscopy (fNIRS) \cite{chao2021fnirs} have been increasingly utilized as inputs for machine learning models, particularly deep learning, in the diagnosis of depression. The use of brain imaging for depression diagnosis provides several advantages, including: 1) Functional brain images provide an objective way to observe and assess structural and functional changes in the brain, especially in neural networks associated with depression such as emotion regulation, reward processing, and cognitive function. 2) Brain functional imaging techniques can help detect depression earlier, especially in individuals whose symptoms have not yet manifested. 3) Functional brain imaging can explore the biological underpinnings of depression and help researchers identify the neurobiological characteristics and mechanisms associated with depression to drive the development of new treatments.



In recent years, fNIRS-based depression diagnosis methods have gained significant attention due to its noninvasive, safe, cost-effective, and portable nature as an optical brain imaging technique \cite{chao2021fnirs}. Currently, two types of fNIRS data are used for depression diagnosis: resting-state fNIRS (RS-fNIRS)~\cite{lu2010use} and verbal fluency task fNIRS (VFT-fNIRS) \cite{ho2020diagnostic}. RS-fNIRS are recorded when subjects sit in a chair with their eyes closed. Subjects are instructed to remain still without thinking about anything. 
VFT-fNIRS comprise a pre-task period (silent period), a task period, and a post-task period (other silent period). Participants are instructed to maintain a fixed gaze at the center of the screen, counting repeatedly from 1 to 16 in the silent period and from 1 to 32 in the other silent period. During the task, participants are asked to verbally generate as many words as possible that begin with a specific character. By recording changes in oxyhemoglobin (HbO) and deoxygenated (HbR) hemoglobin in different brain channels over time during each period, multichannel signal acquisition devices are responsible for imaging several brain channels using near-infrared light sources. After preprocessing the data of the fNIRS, we would obtain a temporal series for each channel of the brain, which is called temporal features (TFs)~\cite{yu2022gnn}. Functional connectivity (FC)~\cite{friston1994functional} employs signals recorded in different regions of the brain to calculate a certain index that reflects the strength of the relationship between different channels of the brain, which is called spatial features (SFs)~\cite{yu2022gnn}.\par

        Earlier studies based on traditional machine learning methods typically considered either spatial features (SFs) or temporal features (TFs), but not both. To address this limitation, Yu et al.~\cite{yu2022gnn} proposed a GNN-based approach that leverages both types of features by treating edges as SFs and nodes as TFs. Although their method achieved a high classification accuracy of 0.85, it suffered from a relatively low F1 score (0.57), which is a critical metric in scenarios with imbalanced classes, such as disease diagnosis. Through an in-depth investigation and reproduction of existing methods, we identify three main shortcomings in previous research: 
        1) Most existing methods that are trained on small datasets (less than 100 patients' fNIRS data~\cite{yu2022gnn,zhang2024improving}) are highly sensitive to class imbalances. This sensitivity often results in suboptimal F1 scores and/or lower recall rates. 
        2) In the field of time-series analysis, the transformation from the time domain to the frequency domain is a fundamental technique. However, prior algorithms have often failed to incorporate this critical step, which may limit their ability to capture efficient and interpretable temporal features (TFs).
        3) In modeling VFT data, previous studies did not differentiate between distinct task stages during network design. We posit that each stage may exert varying influences on classification performance—a hypothesis we aim to validate through subsequent data analysis.


       
\begin{figure*}
    \centering
    \includegraphics[width=1\linewidth]{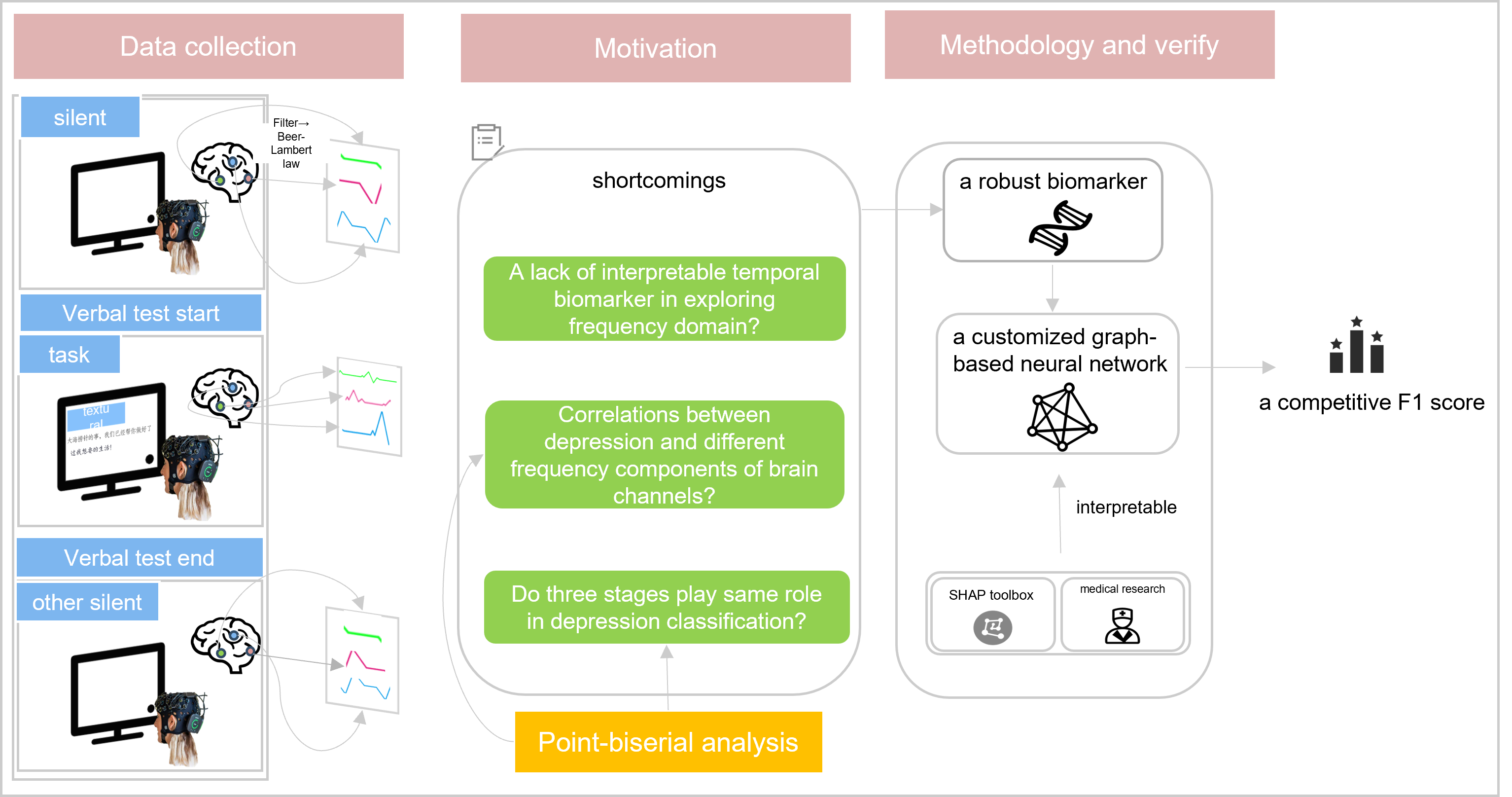}
    \caption{An overview of the paper. Our methodology which consists a robust biomarker and a customized graph-based neural network manages to resolve shortcomings in depression diagnosis task and got a competitive F1 score on our larger dataset.}
    \label{fig:processing_flow}
\end{figure*}

       To overcome these shortcomings, we introduce a robust temporal biomarker for depression diagnosis with a phased TGCN structure to extract fNIRS information sufficiently. 
       First, we built a 10 times larger fNIRS dataset for depression diagnosis to alleviate the issue of class imbalances and enhance the generalization of the diagnosis model. 
       Second, we transfer temporal series to frequency domain to extract  efficient and interpretable temporal features (TFs). 
       statistical analyses such as point-biserial analyses was conducted to verify our motivation. 
       Based on these findings, we propose our novel biomarker and graph based depression classification network which got competitive results among baselines. Finally, a tool named SHAP was utilized to verify that our biomarker is consistent with pre-existing medical discoveries. The processing flow of this paper is in Fig \ref{fig:processing_flow}.
       
In light of the above, the contribution of our work is as follows:
\begin{itemize}
\item In order to solve the shortcoming 1), we decomposed three periods (silent, task, other silent), which means designing independent GCN block modules for each period.
\item In order to solve the shortcoming 2), we presented a novel TF biomarker that contains information from the frequency domain and a novel TFM module that has the ability to fuse the biomarker with SF. To our knowledge, we are the first to consider frequency information in extracting TF features.
\item In order to solve the shortcoming 3), we utilized Point-Biserial Correlation Analyses to analyze correlation between different frequency domains and classification labels. We designed a Frequency Point Biserial Correlation Attention Module (FAM) to reflect our findings. 
\item Our model structure got competitive test performance compared to existing ML methods and deep learning methods in the depression diagnosis task.
\item We utilized a tool called SHAP to vertify the property of the decomposed GCN architecture and FAM.

\end{itemize}
\section{Related Work}
        In this section, we discuss related work. Firstly, we summarize the existing work on the task of depression classification. Secondly, due to the strong ability to extract spatial and temporal information, we discuss some classical graph networks. Finally, we review existing methods of temporal feature extraction in depression recognition by pointing out their limitations and introduce Discrete Fourier Transform (DFT), which is utilized to extract new depression recognition biomarker in our work.
\subsection{Depression Detection}
        Existing research mainly took advantage of data from three aspects to promote diagnose depression--multi-modal data (such as video, text and audio), Electroencephalogram (EEG) and fNIRS. Yang et al.\cite{yang2019multi}emphasized the significant role of linguistic information from the spoken content in automatic depression assessment and proposed a hybrid framework. Ye et al.\cite{ye2021multi}analyzed low-level audio features, deep spectrum features and word vector features and proposed a multi-modal fusion method. Presently, Zhang et al. \cite{zhang2024improving} introduced transfer learning approach for detecting depression in speech, reaching 0.79 F1-score on DAIC‑WOZ dataset \cite{gratch2014distress}. However, multi-modal data is very susceptible to environment factors such as emotion and data acquisition time. Eliminating those noise factors that are not relevant to diagnosis is a though work where EEG and fNIRS data has a natural advantage. Chen et al.\cite{chen2024resting} constructed a resting-state EEG dataset of 579 participants to identify dynamic patterns within the spatial and temporal feature space of nonpsychotic major depression (NPMD), psychotic major depression (PMD) and schizophrenia (SCZ) by utilizing dynamic functional connectivity (DFC) extracted by sliding windows. Zhu et al. \cite{zhu2020classifying} extracted mean and kurtosis of HbO and HbR temporal series as feature and sent these features into an eXtreme Gradient Boosting (XGBoost, XGB\cite{chen2016xgboost}) for classification. Aiming at verbal fluency task (VFT), Yu et al.\cite{yu2017spatio} utilized correlation and coherence of HbO and HbR in different brain channel to emphasized spatial functional connectivity and promoted a GNN based model to better aggregate temporal and spatial information. Apart from that, Zhu  et al.\cite{zhu2023classification} proposed an  adaptive graph neural network (AGNN) to classify children with autism spectrum disorder (ASD) and typical development (TD) with short-term spontaneous hemodynamic fluctuations. However, most dataset of EEG or fNIRS are still very small now. Despite some efforts have used data augmentation to get better evaluation results, they still adopt complex neural networks such as ViT \cite{dosovitskiy2020image} and AlexNet \cite{krizhevsky2012imagenet} which lack a simple and effective model structure in our task. Moreover, some of them \cite{shao2024fnirs} used holdout cross validation strategy but not k-fold cross validation on a small dataset under 100 individuals which increase the influence of accidental factors on experimental result. We team aim to optimize the depression model structure and unify the training and evaluation phases as well.
\subsection{Graph Network}
         Graph networks are neural networks consist of edges and nodes where a node can be an entity or an individual while an edge can represent connectivity of nodes. In depression recognition, for a subject, a node can be regarded as a brain channel\cite{yu2022gnn}. Moreover, some researches proved that the connectivity of depressing brain states are more stronger or weaker than healthy states so it's nature to regard auto-spectrum power density and Pearson's correlation coefficient as edge features to characterize the strength of functional connections in different brain regions\cite{yu2022gnn}. Many work adopted  such GNN methods instead of CNN or recurrent neural network (RNN) running directly on graphs. There are many types of GNNs. GCN\cite{kipf2016semi}, which aggregated features of neighbour nodes to itself according to adjective matrix can be regarded as a variant of CNN. GraphSAGE\cite{hamilton2017inductive} adopted a method of random sampling to achieve a series of path along which nodes' features were aggregated. Graph Attention Network(GAT)\cite{velivckovic2017graph} introduced an invariant function to aggregate related node features with attention mechanism. GraphSAGE and GAT have good abilities of adapting changing graph structures but GCN not. In our task, since the number of brain channels is fixed, GCN stands out because of its relatively simple forward propagation method. A variant of GCN called Diff-pool GCN\cite{ying2018hierarchical} which pools brain regions hierarchically facilitates performance and diff-pool method can be seamlessly inserted into almost any graph neural network\cite{yu2022gnn}. Fan\cite{fan2024identification} proposed DGG (Dynamic graph generation module) to capture dynamic changes of brain functional connectivity which were around stable long-term correlations. However, although the number of brain channels is fixed, the graph can be considered dynamic because each period is different from others in VFT task(e.g. the spatial connectivity of different brain regions during silent period may not be same as its in task period). Thus, regarding each task period as an independent graph to form a graph snapshots could be a better choice. TGCN\cite{zhao2019t} is a temporal graph network which could emphasize the integrity of the graph which applies a GRU or LSTM\cite{graves2012long} to integrate graph embeddings in different VFT periods.
\subsection{Temporal feature extraction}
        In some earlier work, lacking of temporal feature extraction limited the performance of model. They only utilized some statistics in a time series such as mean, std,min,max,kurtosis and skewness) to represent temporal features. Their temporal characteristics neglected temporal changing trends. e.g., it's possible that two different temporal series may share same mean and std but have different temporal curve shapes. In order to overcome the problem, some work utilize full temporal series. Fan\cite{fan2024identification} proposed TCN(Temporal Convolution) module which has two dilated inception convolution (DIC) layers which consisted of multi Conv1d kernel to capture intrinsic temporal correlations. Ma\cite{ma2020distinguishing} referred their VFT task model as AttentionLSTM-InceptionTime (ALSTMIT) , which utilized attention mechanism \cite{vaswani2017attention} to force model learn connectivity of distant brains and used LSTM\cite{graves2012long} to extract temporal information. These methods focused on  time domain and  suffered from the problem of large computation and long-term dependency. Moreover, features in time domain may contain noise that has not been removed during training, which may be harm to accurate depression recognition. Converting time series into frequency domain is benefit to remove those noise. Recently, as a method in signal processing to change data from temporal domain into spectrum domain, Discrete Fourier Transform(DFT) \cite{sundararajan2001discrete} has already been introduced into some deep learning tasks, such as edge extraction\cite{kaur2021fractional}, image denoising \cite{komatsu20173} and autonomous driving trajectory prediction\cite{wong2022view} . It decomposes series inputs in temporal domain into a series of sinusoids with different amplitudes and phases on different frequencies which can present more complex and elusive features which may not be expressed in temporal domain. In this paper, Discrete Fourier transform (DFT) is used to extract spectrum features of  HbO and HbR concentration  in different task stages as temporal features (TFs). Although we are not the first to use DFT in depression recognization, previous work either used DFT to extract cross-spectrum power density as spatial features(SFs)\cite{yu2022gnn} or only utilized finite statistics of fixed bands for all subjects to represent  frequency domain feature which may neglect individual difference and fine-grained feature of dynamic changing time series\cite{wang2021depression}.Distinct from them, our novel biomarker and model are capable to  utilize more specific bands and  frequencies flexibly according to their frequency magnitudes.
\begin{figure*}
    \centering
    \includegraphics[width=0.75\linewidth]{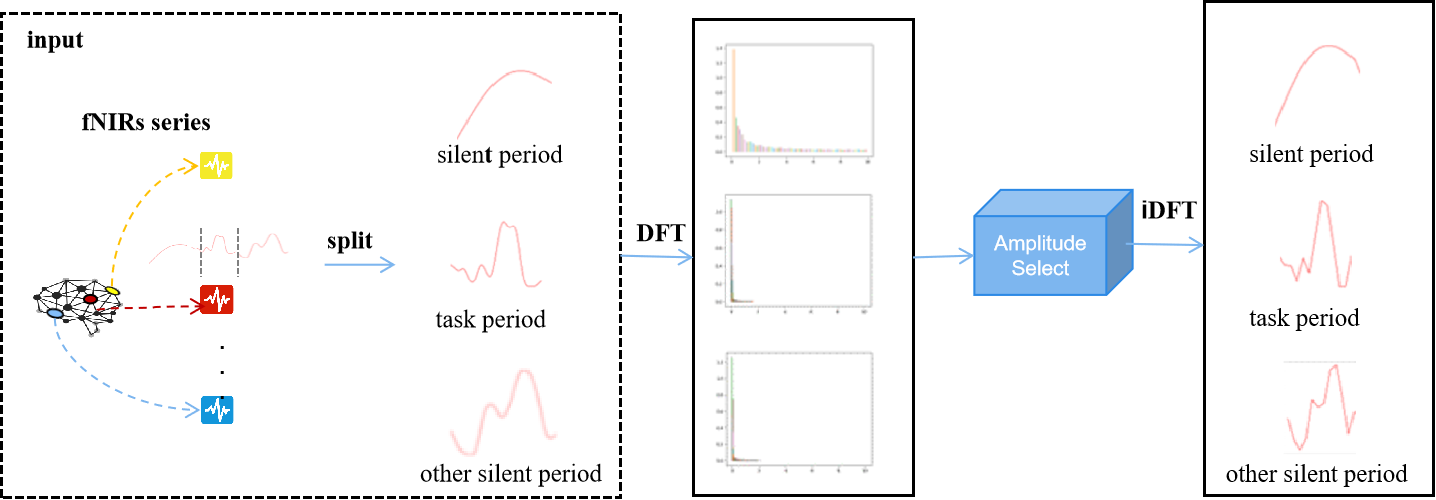}
    \caption{Diagram of DFT and iDFT in our work. After amplitude selection and iDFT, new temporal brain channel series reserve basic trends of origin temporal series. The middle topk frequency domain compoments can be used to extract our biomarkers.}
    \label{fig:dft}
\end{figure*}
\section{Data Preparation}
        The dataset used in this article was collected from the Renmin Hospital of Wuhan University and volunteers were tested under experimental conditions. This section presents the details of data collection, processing, and feature extraction.
\subsection{Dataset}
        Our VFT-fNIRS measurement was conducted with a 53-channel infrared NIR spectrometer system(BS-3000, Znion Technology Company, Wuhan, China). The absorption of near-infrared light in the prefrontal cortex at wavelengths of 695 and 830 nm was assessed using a sample rate of 20Hz. Our VFT contains a silent period of 30 seconds, a task period of 60 seconds, and another silent period of 60 seconds. The origin depression recognition dataset recruits 10,86 subjects with 351 patients and 735 normal individuals.\par
        After collecting data, we used the Homer2 software package\cite{huppert2009homer}, a Matlab-based toolbox to preprocess VFT-fNIRS. Firstly, we examined our data to have eliminated artifacts such as motion-related noises which were addressed using cubic interpolation. Then, the option intensity was transformed into the optical density. The resulting interpolation was subtracted from the original signal, after which the denoised segment was obtained (3s for tMask, 0.5s for tMotion, AMPthresh=5 and STDEVthresh=20)\cite{scholkmann2010detect}. The data were then filtered using a bandpass filter (0.01-0.1Hz) to remove cardiac interference and baseline drift, and then to isolate spontaneous neural activity\cite{santosa2013noise}. We transform the optical density into the relative concentration values of HbO and HbR following the revised Beer-Lambert law\cite{cope1988system}. To ensure steady state conditions, the first 10 seconds and the last 2 minutes of data were excluded. More details of our pre-processed entire dataset subjects(real-world dataset) are in the Tab\ref{tab:real_world_subjects_infor}.
\begin{table}[ht!]
  \begin{center}
    \caption{subject information(real\_world)}
    \label{tab:real_world_subjects_infor}  \begin{tabular}{c|c c }
    \toprule[2pt]
      \textbf{} & {Depression}& {Normal}\\
      \hline
      Number& 351& 735\\
      Gender& 242females,109males& 247females, 488males\\
      Age mean& 15.19& 12.76\\
      Education length mean& 8.46& 6.41\\
    \bottomrule[2pt]
    \end{tabular}
  \end{center}
\end{table}
        Moreover, in order to make the selected subjects comparable in clinical characteristics, we performed propensity matching (PSM) \cite{kane2020propensity} of all patients to obtain a new dataset to test the performance of our model on a dataset that meets medical statistical requirements. The PSM dataset contains 420 participants with 210 patients and 210 normal individuals. The whole research was assisted by many professional mental doctors who we want to express great thank to.
\subsection{Feature extraction}
        Our features contain two parts: temporal features (TF) and spatial features (SF) which we will discuss how to extract them in this section. 
        Before we extract TF and SF, we should preprocess the raw data. Since our origin data has a magnitude about 1e-7, our first step is normalizing origin data along temporal axis for each period to prevent gradient vanishing and gradient explosion in the following training. We denote $T = [t_{1},t_{2},...,t_{N}]$ as an N-length HbO (HbR) sequence of any period and $\mu$, $\sigma$ are sample mean and sample standard deviation (std) of the corresponding period respectively. The normalized data for time i $t'_{i}$ is calculated as (1).
\begin{equation}
t'_{i} = \frac{t_{i} - \mu}{\sigma}
\end{equation}
        Our temporal features (TF) also consist of two parts : Origin Temporal Features (OTF) and Dynamic Temporal Features(DTF). For OTFs, like most earlier work, we firstly calculate 6 statistics(mean, std, min, max, kurtosis and skewness) of each VFT period normalized data respectively, written as $O\in R^{c*6}$. We noticed different statistics varied greatly so we made another normalizing operator along brain channel axis. This normalizing made each statistic contributes equally at the start of training. OTF can be regarded as scale of temporal series. For DTF, we make Discrete Fourier Transform on temporal series of each period as (2). e.g. for the task period, we have $c*l_{task}$ temporal points where c means the number of brain channels and $l_{task}$ means length of task periods. After Discrete Fourier Transform operation along $l_{task}$ axis, we get a spectrum feature $D_{t}\in R^{c*l_{task}*2}$ where '2' represents amplitude $A$ as (3) and phase $\Phi$ as (4) in spectrum domain. $A$ and $\Phi$ are calculated as follows where k means k-th frequency partials.\par
\begin{equation}
T[k] = \sum_{i=0}^n x[i]cos(2\pi\frac{ki}{N})+j\sum_{i=0}^n x[i]sin(2\pi\frac{ki}{N})
\end{equation}
\begin{equation}
A[k] = \Vert T[k] \Vert
\end{equation}
\begin{equation}
\phi[k] = angle(T[k])\
\end{equation}
        To remove potential noises in DTFs, we select those frequencies that have topk amplitudes like \cite{wu2022timesnet}. k is a hyper-parameter to influence the degree of noise removal. That's because when we conduct inverse Discrete Fourier Transform, the fewer frequencies we select, the coarser temporal series we will get. Fig \ref{fig:dft} demonstrates the process. We use inverse Discrete Fourier Transform to reform key spectrum into temporal. Result illustrates that it's feasible to select few frequencies to restore such temporal series preserving origin series curve shape. After Discrete Fourier Transform processing and amplitude select, we got dynamic temporal features $D \in R^{c*k*2}$ where k represents the number of sinusoid partials.
        We also recorded the frequency indices with topk amplitude as $F \in R^{c*k}$. We extract our  biomarker utilizing $\{D, F\}$ with a TFM module which will be discussed in the next section for each period independently. The biomarker can be regarded as fine-grained feature such as concentration curve shape based on temporal series' scale (OTF).\par
        Like \cite{yu2022gnn}, Our SF $S\in R^{c*c*2}$continues to utilize Pearson's correlation (corr) coefficient and coherence (cohe) to represent spatial connection of different brain regions for each period respectively. Given two brain channel sequences $X$ and $Y$, their correlation and coherence are as (5)-(6), where $S_{XY}$ refer to cross-spectrum power density and $S_{XX}$ refer to auto-spectrum power density:
\begin{equation}
Cor(X,Y) = \dfrac{\sum_{i}^{t}(X^i - \overline{X})(Y^i - \overline{Y})}{\sqrt{\sum_{i}^{t}(X^i - \overline{X})}\sqrt{\sum_{i}^{t}(Y^i - \overline{Y})}}
\end{equation}
\begin{equation}
    Cohe_{XY}{(f)} = \dfrac{|S_{XY}(f)|^2}{S_{XX}(f)S_{YY}(f)}
\end{equation}
Correlation reflects linear correlation degree within temporal domain and coherence reflects interrelationship between two signal within spectrum domain via calculating their cross-spectrum power density and auto-spectrum power density. Earlier work has proved it's more effective to leverage average correlation and coherence of subjects in training dataset as a prior information instead of using individual correlation and coherence\cite{yu2022gnn}. \par
\begin{figure*}
    \centering
    \includegraphics[width=1\linewidth]{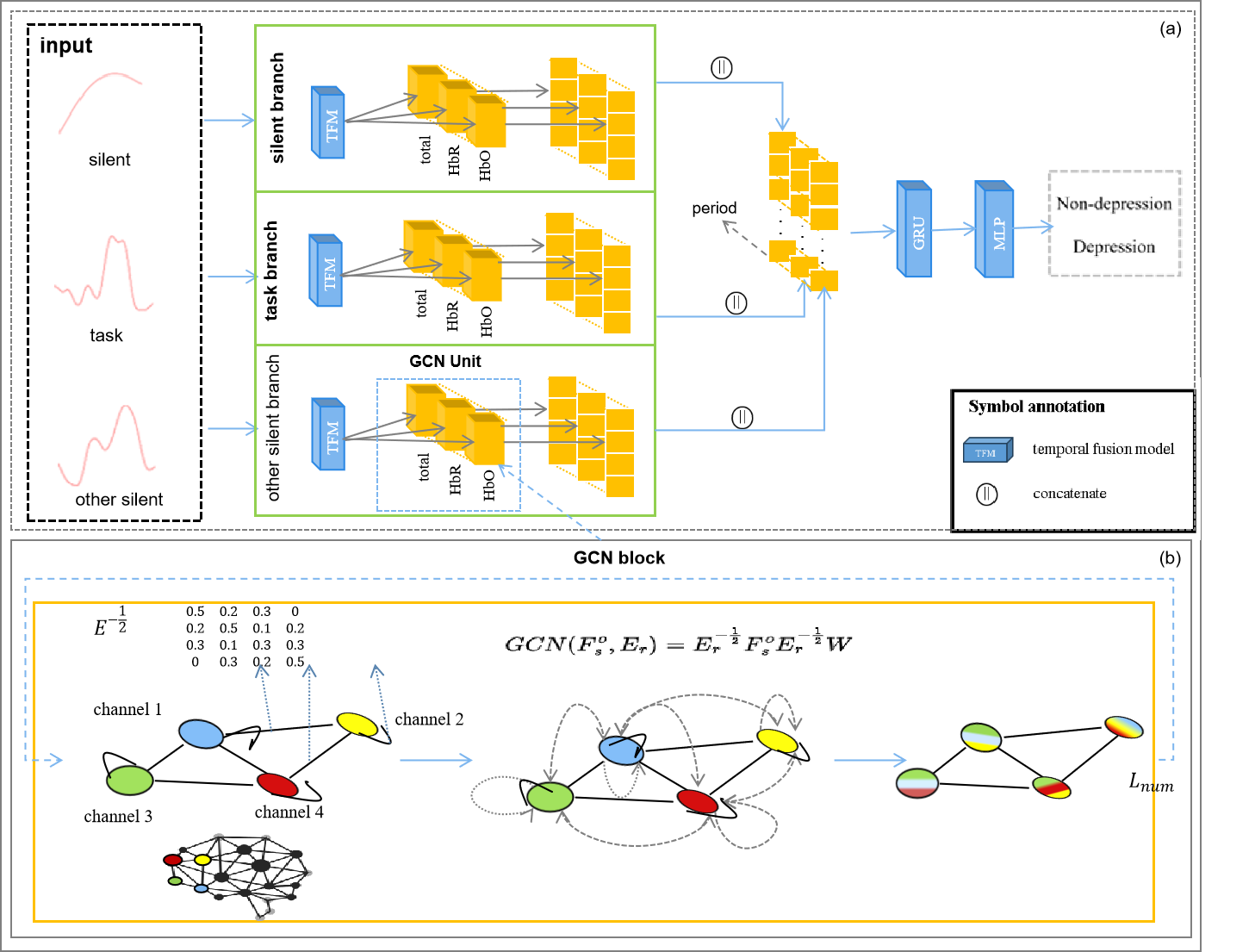}
    \caption{(a) An overview of our TGCN based model. Three independent TGCN modules deal with HbO, HbR and their total concentration representively. (b)  Details of GCN block which was used to aggregate information from neighbor brain channels in this work.}
    \label{fig:total_structure}
\end{figure*}
\subsection{Data Driven Motivation}
\begin{figure*}[h]
    \centering
    \includegraphics[width=0.75\linewidth]{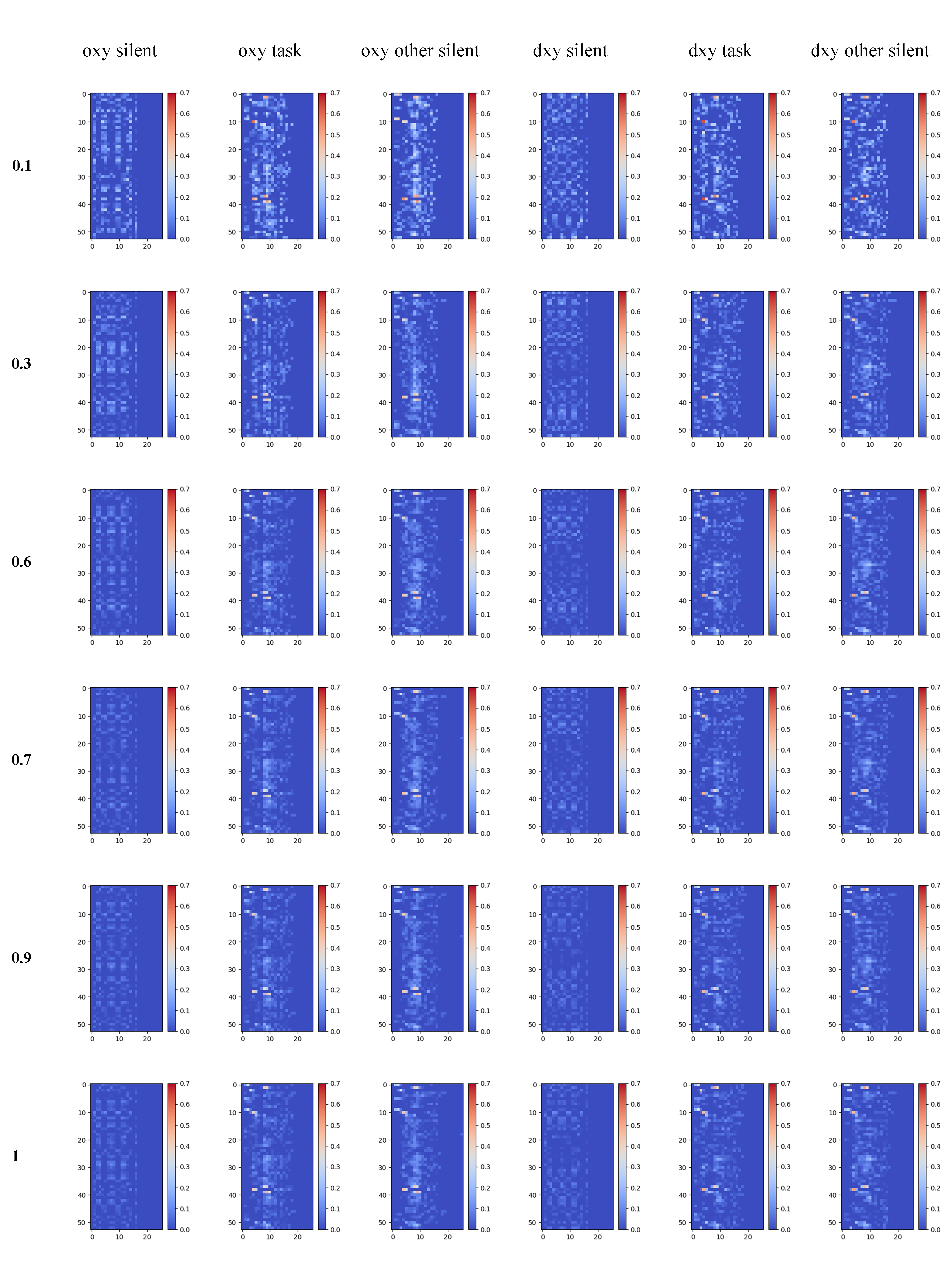}
    \caption{Consistency vertification of our novel biomarker. Y-axis represents brain channels and x-axis represents frequencies' index, where a larger x coordinate represents sinusoid with higher frequency. Color means relative size of $|r|$.We found that frequency component of specific brain regions was strongly associated with depression. }
    \label{fig:point-biseral}
\end{figure*}

        Before we illustrate our methodology,  we analyze the correlation between the k top amplitudes and the ground truth label. For simplicity, we selected frequencies with highest k amplitudes to represent biomarkers for each brain channel of each period. Point-biserial analyses manages to measure correlation between a continuous random variant X and a binary random variable $Y$. We formulate hypotheses $H_{1}$: The biomarker feature $T_{i}$ has a significant impact on classification results $Y$. We construct hypothesis test statistics $r$ as in (7).
\begin{equation}
r(T_{i}, Y) = \frac{T_{i}^{n} - T_{i}^{p}}{\sigma} \times \sqrt{np}
\end{equation}
      where $T_{i}^{n}$, $T_{i}^{p}$denotes average feature $i$ of normal individuals and patients respectively, $\sigma$ denotes feature std of all dataset, n and p denote the the number of normal and patient individuals respectively. The larger $|r|$ indicates a stronger correlation between the feature and the label. As mentioned above, we calculate the highest k amplitudes for each brain channel as DTF $\in R^{c*k}$. To access whether our DTF are consistent across different sample sizes, we calculate $|r|$ on randomly selected proportions of the data like \cite{chen2024resting} in Fig. \ref{fig:point-biseral} where the first column represents the proportion of the selected training dataset. We find that: \par
\begin{itemize}
    \item The silent period, task period, and other silent periods have distinct impacts on classification results (max $|r|$ of the silent stage $<$ 0.2 and max $|r|$ of the task stage and other silent stage $>$ 0.4), indicating that we should \textbf{model these three phases separately}. \par
\end{itemize}
\begin{itemize}
    \item Higher frequency means less correlation to classification results (the right area, which represents high frequencies, is bluer than the left area, which represents low effects on classification). This finding preliminarily supports the rationality of \textbf{amplitude selection}. \par
\end{itemize}
\begin{itemize}
    \item Features with significant impacts on the classification results (red points) are mainly concentrated in a limited number of sinusoids in only a few brain channels (0-2, 9-11, 37-40 channels) in task and other silent stages. This finding might inspire us to \textbf{design an attention module} to distinguish the importance of different channels and frequencies. \par
\end{itemize}
    Our motivation and methodology are all based on these findings.
\section{Methodology}
       Since we have TF and SF in the feature extraction section, we will discuss how to fuse them in this section.
       In Section A, we will illustrate our model in general, but we will not focus on the concrete TFM module whose details will be shown in Section B. In Section C, we will discuss how to design the attention module FAM.
\subsection{Phased Network Structure}
        According to shortcoming and finding 1), as in Fig. \ref{fig:total_structure}, our model consists of three the same structures for each period. Firstly, we slice the temporal series at the junction of three periods. Then we sent temporal series into our carefully designed temporal fusion module (TFM) to learn node (brain channel) embeddings for each period independently. We have three GCN groups for HbO, HbR, and their total concentrations. Correlation and coherence can be regarded both as adjacency relations and edge weights of GCN, so there are two independent GCNs for each group. Thus, we designed 2*3 = 6 same and parallel GCNs to fuse TF with SF for each period. \par
        GCN is a convolutional neural network based on a fixed graph structure. In GCN, each node has an embedding feature that can be updated iteratively using its adjacencies. Firstly, for each node, GCN uses aggregate functions which satisfy permutation invariant (e.g. sum, average, normalize) to aggregate neighbors' features to itself. Then, the aggregated feature for each node will go through an MLP to enhance the nonlinear expression ability of the network. Finally, each node will use the new feature to update itself. In practical use, people often repeat the aforementioned process until the node features are stable. Suppose $n$ denotes the number of nodes, $E_{r}^{-\frac{1}{2}} \in R^{n*n}$ denotes the normalized adjacency matrix, $W \in d_k * d_{k}$ denotes the trainable weights of the GCN, $F_{s}^{o} \in R^{n * d_{k}}$ denotes the characteristics of substance $o$ of period $s$ before updating. Then the updated features by GCN can be regarded as (6).
\begin{equation}
GCN(F_{s}^{o}, E_{r}) = E_{r}^{-\frac{1}{2}}F_{s}^{o} E_{r}^{-\frac{1}{2}}W
\end{equation}
        In our task, as Alg. \ref{alg:GCN}, nodes represent brain channels, SF can be considered as E, and features that pass through TFM can be considered as node characteristics $F$. After aggregation, we use another MLP to update the aggregated features. We repeat the process $L_{num}$ times to obtain the final stable characteristics $L_{1}$. Now we have $L_{1}$ for silent, task, and other silent periods, so we can use a 3 layer GRU module (Gate Recurrent Unit) \cite{chung2014empirical} to process stage series. Finally, we use a softmax layer to obtain the output probabilities for binary classification. \par   
\begin{algorithm}[!pht]
    \label{alg:GCN}
    \renewcommand{\algorithmicrequire}{\textbf{Input:}}
	\renewcommand{\algorithmicensure}{\textbf{Output:}}
	\caption{forward propagation process for our model}
    \label{power}
    \begin{algorithmic}[1] 
        \REQUIRE ; fNIRS time series data $T$, spatial feature $E$, GCN layer num $L_{num}$
	    \ENSURE Depression probability $\hat{y}$; 
        
        \STATE $S \xrightarrow{}${$silent, task,other silent$}    \%split temporal series
        \STATE $O \xrightarrow{}${$HbR$,  $HbO$, $total$}    \% substances 
        \STATE $R\xrightarrow{}${$corr, cohe$}   \% correlation and coherence
         \STATE $L_{2} = [\,] $
        \FOR{each $s$ in $S$ in parallel}
        \STATE $L_{1} = [\,] $
        \FOR{each $o$ in $O$ in parallel}
            \STATE $F_{s}^{o} = $ $TFM(T_{s}^{o} )$  $\in R^{53*d_k}$  \%temporal embedding
           \FOR{each $r$ in $R$ in parallel} 
                \FORALL {$i=1,2,\cdots, L_{num}$}
                    \STATE  $F_{s}^{o} = MLP(GCN(F_{s}^{o}, E_{r}))$    \% $E_{r} \in R^{53 * 53}$
                \ENDFOR
            \STATE $L_{1} \xleftarrow{} f_{s}^{o}$
           \ENDFOR
        \ENDFOR
        \STATE  $L_{1} \in R^{3*2*53*d_k} \xrightarrow{} R^{53 * (6*d_k)}$  \%concatenate
        \STATE $L_{2} \xleftarrow{} L_{1}$
        \ENDFOR
        \STATE $L_{2} \in R^{3*53*(6*d_k)}$     \%concatenate
        \STATE $y_{pred}=softmax(MLP(GRU(L_{2}))) $
        \STATE \textbf{return} $\hat{y}  \in R^{53*2}$.
    \end{algorithmic}
\end{algorithm}
———————————————
\subsection{Temporal Fusion Module Considering Frequency Domain}
\begin{figure*}[tbp]
    \centering
    \includegraphics[width=1\linewidth]{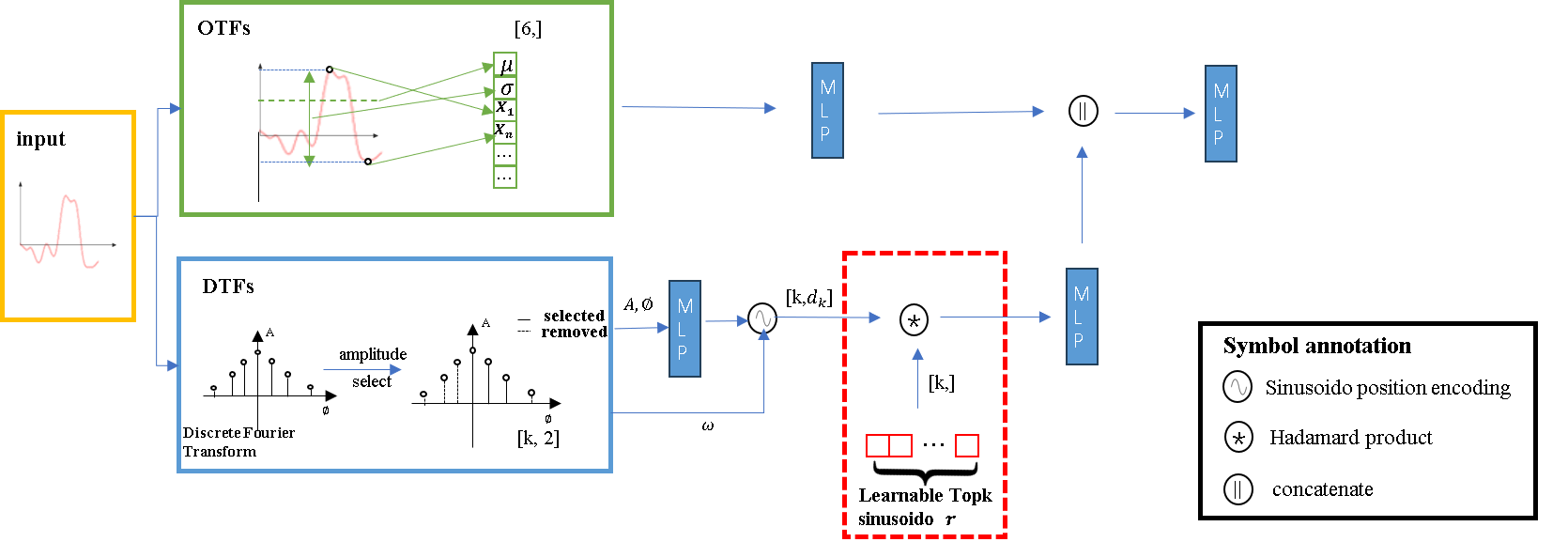}
    \caption{Temporal fusion model. Green block means Original Temporal Feature which contains six statistics of FNIRS. Blue block means our Dynamic Temporal Feature under frequency domain. Red dot block represents alternative FAM module to enhance model performance.}
    \label{fig:temporal fusion module}
\end{figure*}
        According to shortcoming and finding 2), in this section, we highlight our new biomarker and a carefully module TFM tailor made for it.
        For each stage, we carefully designed a temporal fusion module to better fuse OTF and DTF together. For simplify, we only consider a single brain channel. Let $T = [t_{1},t_{2},...,t_{n}]\in R^{n}$ be a normalized $n$-length temporal series for a brain channel. We first calculated mean, std, min, max, kurtosis and skewness of $T$ as primeval OTF origin information. Then an MLP is used to capture OTF feature $f_{o}$ as (6).\par
\begin{equation}
f_{o}  = MLP([\mu,\sigma, T_{1},T_{n},Kurt,Skew])
\end{equation}
        For DTF, we first change them from temporal domain to spectrum domain as (7):
\begin{equation}
S = DFT(T)\in C^{n}
\end{equation}
       we got primeval biomarker via calculating amplitudes and phases of all $n$ sinusoids. After that , we sort sinusuids as amplitudes and select the top k max amplitudes $D_{k}$ and record their frequencies index as $F =[\omega_{1},\omega_{2},...,\omega_{k}] \in I^{k}$. Our new biomarker is defined as $\{D, F\}$. Bands' sequence $F$ of different subjects can be distinct and choosing same bands' sequence for all subjects can neglect individual difference. An MLP is used to extract biomarker feature $f_{d}$ here as (11)
\begin{equation}
f_{d} = MLP([ { \Vert S_{k} \Vert}, sin(arg S_{k}), cos(argS_{k})]) \in R^{k*d_{k}}
\end{equation}
        Please notice that here $w_{i}$ may not equal to $i$. Thus, a position encoding module PE \cite{vaswani2017attention}$\in R^{k*d_k}$ can be added  to $f_{d}$ in order to enhance frequency location information as (12)-(14).
\begin{equation}
PE(\omega_{i}, 2j) = sin(\frac{\omega_{i}}{10000^{\frac{2j}{d_{k}}}})
\end{equation}
\begin{equation}
PE(\omega_{i}, 2j+1) = cos(\frac{\omega_{i}}{10000^{\frac{2j+1}{d_{k}}}})
\end{equation}
\begin{equation}
f_{d} = f_{d} + PE \in R^{k*d_k}
\end{equation}
       ,where $2j$, $2j+1$ denote even feature dims and odd feature dims respectively.
\subsection{Frequency point biserial correlation attention module}
       According to shortcoming and finding 3), only few brain channels have large affects on labels and not all frequencies in these brain regions have large affects on labels. Thus, we suggest to add an alternative frequency  point biserial correlation attention module (FAM, red dotted box in \ref{fig:temporal fusion module}). Thanks to the consistent of $r$ statistics in point biserial correlation analyses before, we can use coefficient of training dataset in \ref{fig:point-biseral} $W \in R^{k}$ as initial weights of DTF. (Be careful we can only use training dataset to calculate $W$ otherwise data breaches will occur!) A larger $|r|$ indicate the frequency may have more effect to classification and this will force model pay more attention on specific frequency domain by conducting Hadamard product between $W$ and $f_{d}$. $W$ are learnable parameters in (15).
\begin{equation}
f_{d} = MLP(Hadamard(W,f_{d})) 
\end{equation}
        Finally, we concat OTF features with DTF features and use an MLP to get final temporal features $f$ which is input of GCNs as (16).
\begin{equation}
f = MLP(f_{o} \Vert f_{d})
\end{equation}
\section{Experiment}
\subsection{Experiment Setup}
        First of all, we competed our model with several existing methods which contain traditional machine learning methods and a classical SOTA method of depression recognition based on GNN under both real-world and PSM circumstance. Then for the ablation studies, we selected top1 to top10 amplitudes to generate our new biomarker to compare their results. Furthermore, we tried to remove alternative frequency  point biserial correlation attention module (FAM) to investigate its influence.\par
        Our MLP block consists of a linear1d and activation function of ELU. We used Cross Entropy loss as loss function and focal loss($\gamma = 1.5, \omega = 0.5$) was used as a trick as well.  Moreover, we used L2-regularization to avoid overfitting. We used 1e-3 as learning rate and Adam as optimizer. A MultiStepLR with milestone$[30,60,90]$ and $\gamma=0.5$ was used as learning rate scheduler. Despite our dataset is larger than others, 4-fold cross validation was used to guarantee more convincing results. When splitting dataset in k-fold cross validation, stratified shuffle split was used to balance the rate of normal and patient in training dataset and validation dataset. We fixed random seed as 2024 to make sure  baseline and our own model use congruous  data. We noticed that our dataset is imbalanced (735:267), so apart from binary accuracy, we also introduced micro precision, micro recall, micro F1-score as evaluation metrics. Mean evaluation metrics of 4 test folds were used to represent final results and we selected the best epoch with highest F1-score to represent model performance.\par
        The experiments were all carried out with one NVIDIA RTX 4090 GPU  implemented by Python 3.8.18. We used sklearn and Pytorch\cite{paszke2019pytorch} to reproduce baseline models and conduct our experiments.
\subsection{Experiment Results}
         Before we start our experiment, we noticed different methods use private dataset with distinct sizes and category proportions. In order to unify the results of the experiment, we randomly sample 96 subjects from our dataset with same size and category proportions to previous work\cite{yu2022gnn}\cite{shao2024fnirs}. We regarded it as baseline simulation dataset and got results on it as Tab\ref{tab:reproducing_small_dataset}. The baseline GCN model that we reproduced(whose results were shown in penultimate row) has almost same accuracy and F1-score as in paper, indicating we successfully reproduced baseline model. We tested baseline model on our larger dataset(1086 samples) and found that Accuracy is 3\% lower than  results on smaller dataset(96 samples). However, those metrics which pay more attention on imbalanced data proportion such as Precision, Recall, F1-score are 14.3\%, 19.4\%,23.9\% promoted respectively. As our dataset is imbalanced, this result reinforce the necessity of introducing these metrics.\par
\begin{table}[ht!]
  \begin{center}
    \caption{Reproduce of baseline model(bold is best).}
    \label{tab:reproducing_small_dataset}  \begin{tabular}{c|c c c c }
    \toprule[2pt]
      {Model} &{Acc.$\uparrow$}& {Precision$\uparrow$}& {Recall$\uparrow$}&{F1$\uparrow$}\\
    \hline
 LeNet(96)& 0.762& 0.637& 0.662&0.646\\
 CNN-GRU(96)& 0.667& 0.389& 0.412&0.4\\
 ViT(96)& 0.714& 0.395& 0.441&0.417\\
      \hline
      {Baseline GCN(96)}&$\textbf{0.854}$& $0.7$&$0.488$&$0.563$\\
      Reproduced GCN(96)&$0.833$& $0.595$&$0.605$&$0.555$\\
      Reproduced GCN(1086)&$0.8075$ & \textbf{$\textbf{0.685}$}&$\textbf{0.735}$&$\textbf{0.705}$\\
    \bottomrule[2pt]
    \end{tabular}
  \end{center}
\end{table}
        The tab \ref{tab:PSM_metrics} illustrates the results of the baselines and our model in a real-world data set. Notice that there is a trade-off relation between Precision and Recall. The F1 score is a combination of both. Among baselines, RF using TFs achieved the highest precision of 0.88 but with a lowest recall of 0.55 and the second lowest F1 score of 0.67. SVM using TF achieved the best F1 score of 0.749 and the best accuracy of 0.838. Specifically, for LR, KNN, and SVM\cite{song2014automatic}, TF works better than SF. However, SF works better than TF for RF\cite{zhu2020classifying}.  Surprisingly, the existing SOTA GNN-based model\cite{yu2022gnn} only achieved an F1 score of 0.75 (11\% lower than ours) on our large-scale dataset, although it claimed to fuse temporal and spatial information. However, utilizing an alternative FAM module, our GNN-based model achieved the highest accuracy, F1 score of 0.864, 0.833, and the second highest precision, recall.\par
\begin{figure}
    \centering
    \includegraphics[width=1\linewidth]{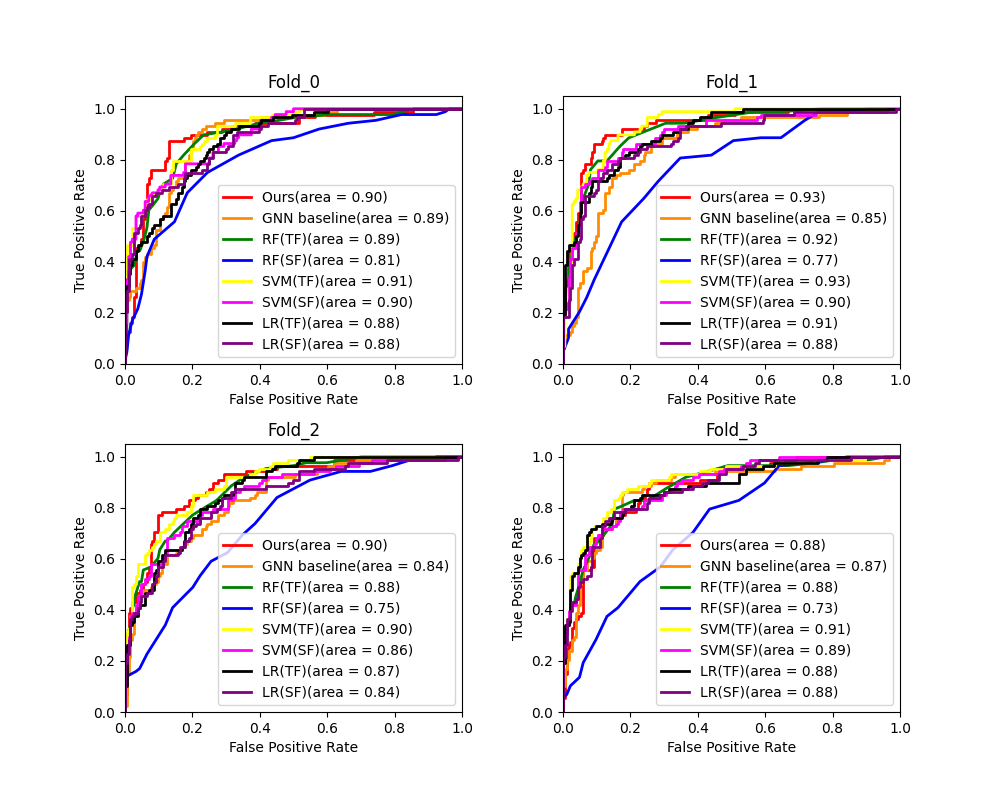}
    \caption{ROC curve of real world dataset where red curve means ours. We adopt 4-fold cross- validation. Compare with other baselines, the curves of our model in fold 0, 1, 2 are more closer to (0, 1) point, which shows the superiority of our model in real-world scenarios..}
    \label{fig:real_world_AUC}
\end{figure}
\begin{figure}
    \centering
    \includegraphics[width=1\linewidth]{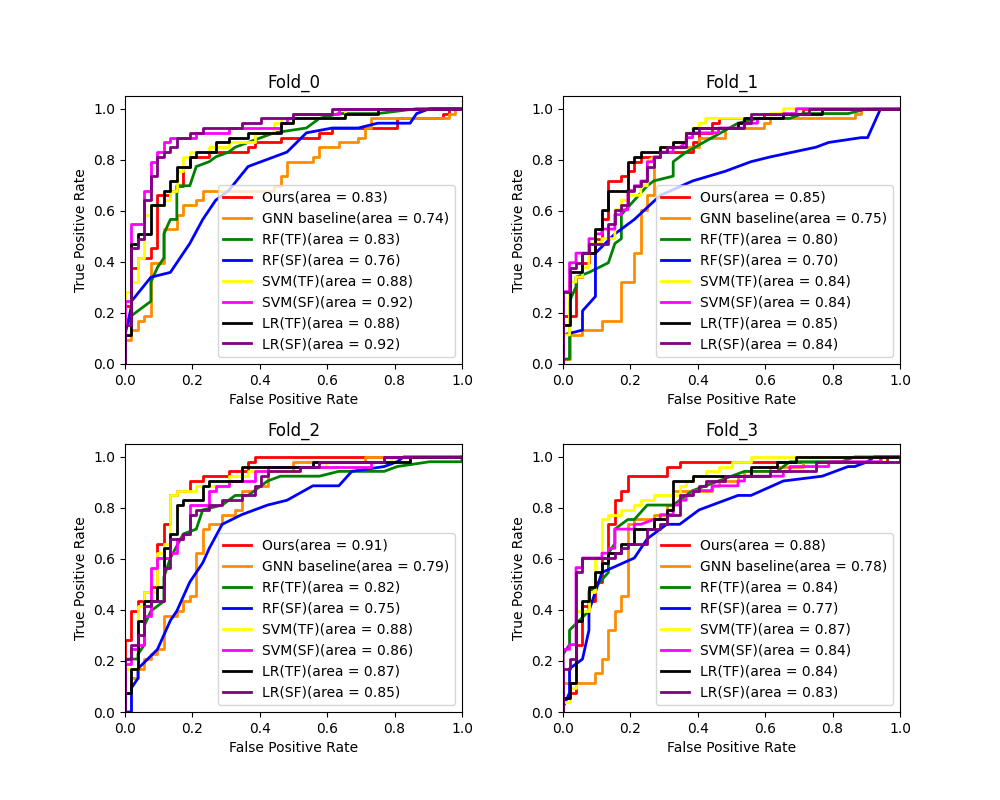}
    \caption{ROC curve of PSM dataset where red curve means ours. We adopt 4-fold cross- validation. Compared with other baselines, the curves of our model in fold 1, 2, 3 are more closer to (0, 1) point, which shows the superiority of our model in medical research scenarios.}
    \label{fig:PSM_AUC}
\end{figure}

        The tab \ref{tab:PSM_metrics} shows that our model got the highest precision, precision, F1 score, and AUC of 0.831, 0.823, 0.833, 0.868, which were 3.0\%, 1.7\%,2.8\%,0.3\% higher than those baselines. In addition, we plotted the ROC curves of the PSM dataset in a 4-folder. From Fig. \ref{fig:PSM_AUC}, we could easily find that our model got the largest AUC among all baselines in fold 1, fold 2, and fold 3. Fig. \ref{fig:real_world_AUC} shows the ROC curve of the real-world dataset of baseline models and our TGCN based model. AUC means the area under the ROC curve that is enclosed by the axes. Briefly, a better model should have a larger AUC and ROC curve which is convex toward the (0.0, 1.0) point.\par
\begin{figure}
    \centering
    \includegraphics[width=1\linewidth]{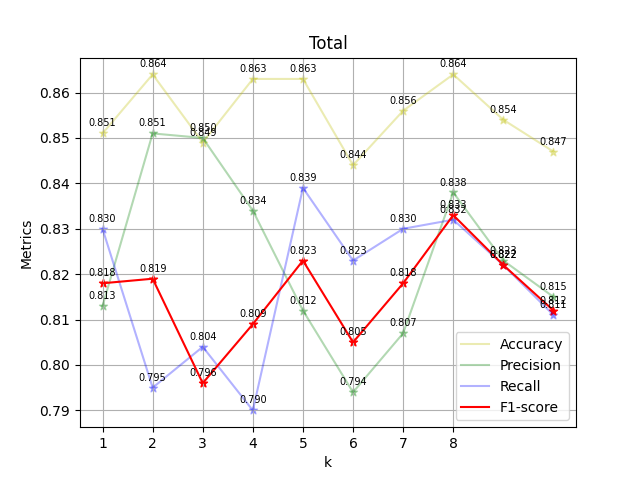}
    \caption{Trend of each metric (Accuracy, Precision, Recall, and F1-score) with increasing of k. Model performance gets promote when k goes from 1 to 8, while all metrics gets decrease when k goes from 8 to 10.}
    \label{fig:k_metric_trending}
\end{figure}

\begin{table*}[ht!]
  \begin{center}
    \caption{Classification results\\Baseline vs Ours(bold is best,underline is second best).}
    \label{tab:PSM_metrics}  \begin{tabular}{c|c c c c | llll}
    \toprule[2pt]
 & \multicolumn{4}{c}{{Real World dataset}} & \multicolumn{4}{c}{{PSM dataset}}\\
      \hline
      {Model} & {Acc.$\uparrow$}& {Precision$\uparrow$}& {Recall$\uparrow$}&{F1$\uparrow$} & {Acc.$\uparrow$}& {Precision$\uparrow$}& {Recall$\uparrow$}&{F1$\uparrow$}\\
      \hline
      LR(TF)& $0.81$ & $0.71$&$0.72$&$0.71$& $0.790$& $0.782$&$0.811$&$0.796$\\
      KNN(TF)& $0.81$ & $0.69$&${0.778}$&$0.73$& $0.705$& $0.638$&$\textbf{0.962}$&$0.767$\\
      RF(TF)& $0.83$& $\textbf{0.88}$&$0.55$&$0.67$& $0.76$& $0.774$&$0.741$&$0.757$\\
      SVM(TF)& $0.838$& $0.76$&$0.74$&${0.749}$& $\underline{0.807}$& $\underline{0.809}$&$0.811$&$\underline{0.81}$\\
 XGB(TF)& $\underline{0.839}$& $0.80$& $0.68$&$0.73$ & $0.776$& $0.771$& $0.797$&$0.783$\\
      LR(SF)& $0.82$& $0.76$&$0.66$&$0.71$ & $0.769$& $0.744$&$0.825$&$0.781$\\
      KNN(SF)& $0.70$& $0.53$&$0.66$&$0.59$ & $0.59$& $0.565$&$\underline{0.849}$&$0.677$\\
      RF(SF)& $0.82$& $0.75$&$0.66$&$0.70$ & $0.69$& $0.699$&$0.684$&$0.69$\\
      SVM(SF)& $0.83$& $0.80$&$0.64$&$0.71$ & $0.771$& $0.738$&$\underline{0.849}$&$0.789$\\
 XGB(SF)& $0.76$& $0.71$& $0.47$&$0.56$ & $0.717$& $0.709$& $0.75$&$0.728$\\
      GNN(TF,SF)& $0.790$& $0.670$&$\textbf{0.878}$&$\underline{0.755}$& $0.748$& $0.710$&$0.815$&$0.753$\\
      \hline
      Ours& ${\textbf{0.864}}$& $\underline{0.838}$&$\underline{0.832}$&$\textbf{0.833}$ & ${\textbf{0.831}}$& $\textbf{0.823}$&$0.846$&$\textbf{0.833}$\\
      \bottomrule[2pt]
    \end{tabular}
  \end{center}
\end{table*}
        The tab \ref{tab:k_ablation_study} and the Fig \ref{fig:k_metric_trending} shows our first ablation study--selection of topk frequency number of DTF. We suppose that using a small k will relieve the computation burden, but neglect some of the brain regions in the Fourier frequency domain that are helpful for classification. However, using a large k will increase the computation burden, but will capture more information from the DTF. Furthermore, we have verified that more information in the frequency domain may not be more conducive to high-precision results through experiments because the F1 score of k = 9 and k = 10 decreases compared to k = 8. We suppose that frequencies with small amplitudes may correspond to noise, which contributes less or even side effects. Thus, it is critical to select the proper k value. We choose k=1 to k=10 and plot the trend of each metric with k. We find that k = 8 got the highest F1 score of 0.833 and the highest accuracy of 0.864 in our dataset.\par
\begin{table}[ht!]
    \begin{center}
    \caption{Classification results with k}
    \label{tab:k_ablation_study}  \begin{tabular}{c|c c c c}
      \toprule[2pt]
      {k}& {Acc.$\uparrow$}& {Precision$\uparrow$}& {Recall$\uparrow$}&{F1$\uparrow$}\\
      \hline
      1& $0.851$& $0.813$&$0.830$&$0.818$\\
      2& $\textbf{0.864}$& $0.849$&$0.795$&$0.819$\\
      3& $0.849$& $\textbf{0.850}$&$0.804$&$0.796$\\
      4& $\underline{0.863}$& $0.834$&$0.790$&$0.809$\\
 5& $\underline{0.863}$& $0.812$& $\textbf{0.839}$&$\underline{0.823}$\\
      6& $0.844$& $0.794$&$0.823$&$0.805$\\
      7& $0.856$& $0.807$&$0.830$&$0.818$\\
      8& $\textbf{0.864}$& $\underline{0.838}$&$\underline{0.832}$&$\textbf{0.833}$\\
 9& $0.854$& $0.823$& $0.822$&$0.822$\\
 10& $0.847$& $0.815$& $0.811$&$0.812$\\
    \bottomrule[2pt]
    \end{tabular}
  \end{center}
\end{table}
        The tab \ref{tab:FAM_ablation_study} shows our ablation study of the FAM module. We selected k = 2, k = 5, and k = 8 to compare the metrics produced by the model with FAM to the metrics produced by the model without FAM. Regardless of k value, model with FAM got higher metric compared with model without FAM, which proved efficiency of our FAM module.
\begin{table}[ht!]
  \begin{center}
    \caption{Ablation study of FAM module}
    \label{tab:FAM_ablation_study}  \begin{tabular}{c c | c c c c} 
    \toprule[2pt]
      {k}&{FAM}& {Acc.$\uparrow$}& {Precision$\uparrow$}& {Recall$\uparrow$}&{F1$\uparrow$}\\
      \hline
 \multirow{2}{*}{2}&-& $0.839$& $0.740$& $\textbf{0.833}$&$0.782$\\ 
 &\checkmark& $\textbf{0.864}$& $\textbf{0.849}$& $0.795$&$\textbf{0.819}$\\ 
      \hline
      \multirow{2}{*}{5}&-& $0.832$& $0.747$&$0.806$&$0.774$\\ 
      &\checkmark& $\textbf{0.863}$& $\textbf{0.812}$&$\textbf{0.839}$&$\textbf{0.823}$\\ 
      \hline
      \multirow{2}{*}{8}&-& $0.843$& $0.774$&$0.802$&$0.786$\\ 
      &\checkmark& $\textbf{0.864}$& $\textbf{0.838}$&$\textbf{0.832}$&$\textbf{0.833}$\\ 
    \bottomrule[2pt]
    \end{tabular}
  \end{center}
\end{table}
\begin{figure*}[tbp]
    \centering
    \includegraphics[width=1\linewidth]{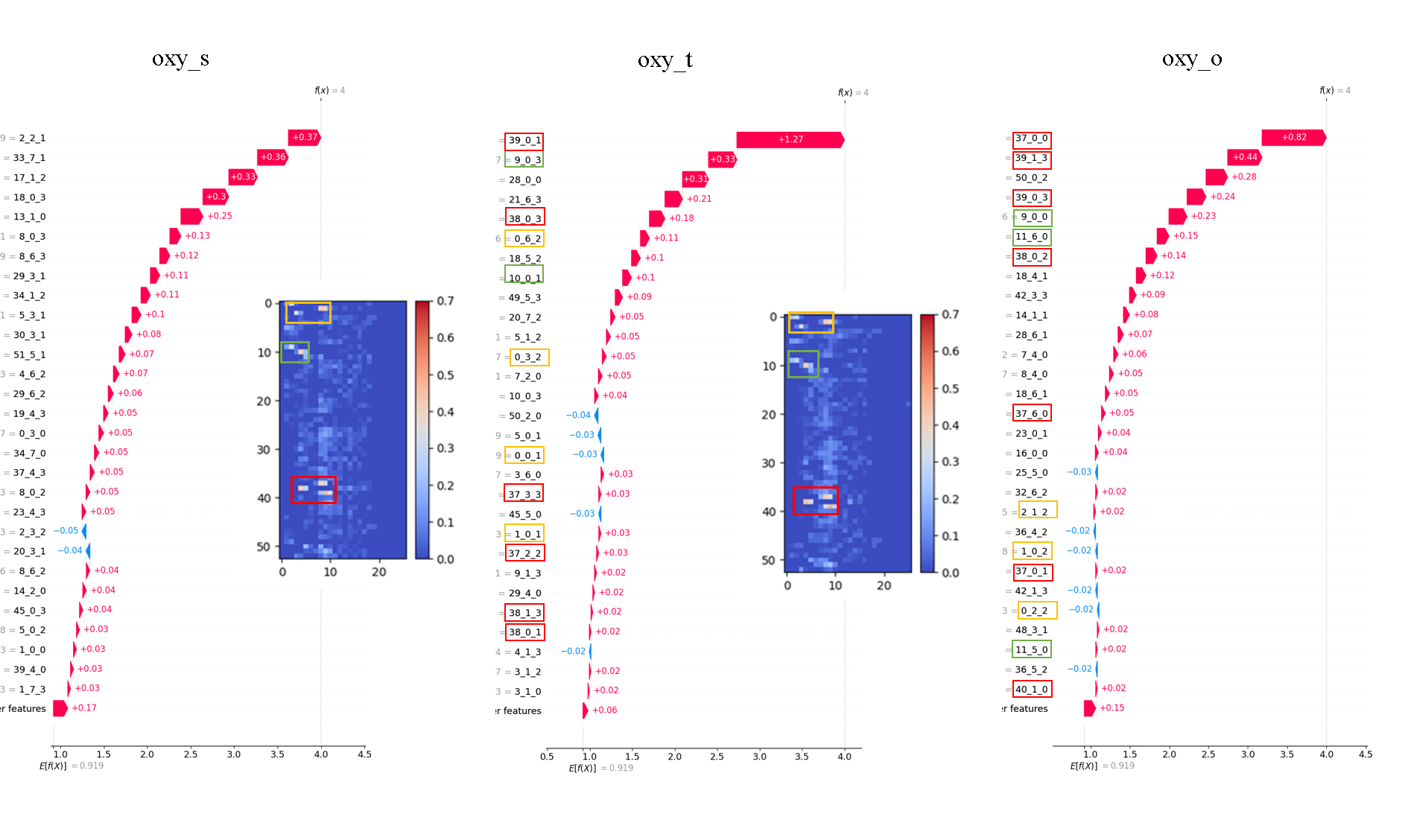}
    \caption{Waterpool picture of HbO. Y-axis denotes feature in format '\textbf{Brain channel}\_\textbf{Topk amplitude indices}\_\textbf{Feature dim}'. X-axis denotes SHAP values. Red represents forward gain and blue represents backward gain. We also use red, green, yellow block to sign significant brain channels during Point-Biserial Correlation Analyses and SHAP analyses to show their correspondence in order to verify interpretable of our approach.}
    \label{fig:HbO_SHAP}
\end{figure*}
\begin{figure*}[tbp]
    \centering
    \includegraphics[width=1\linewidth]{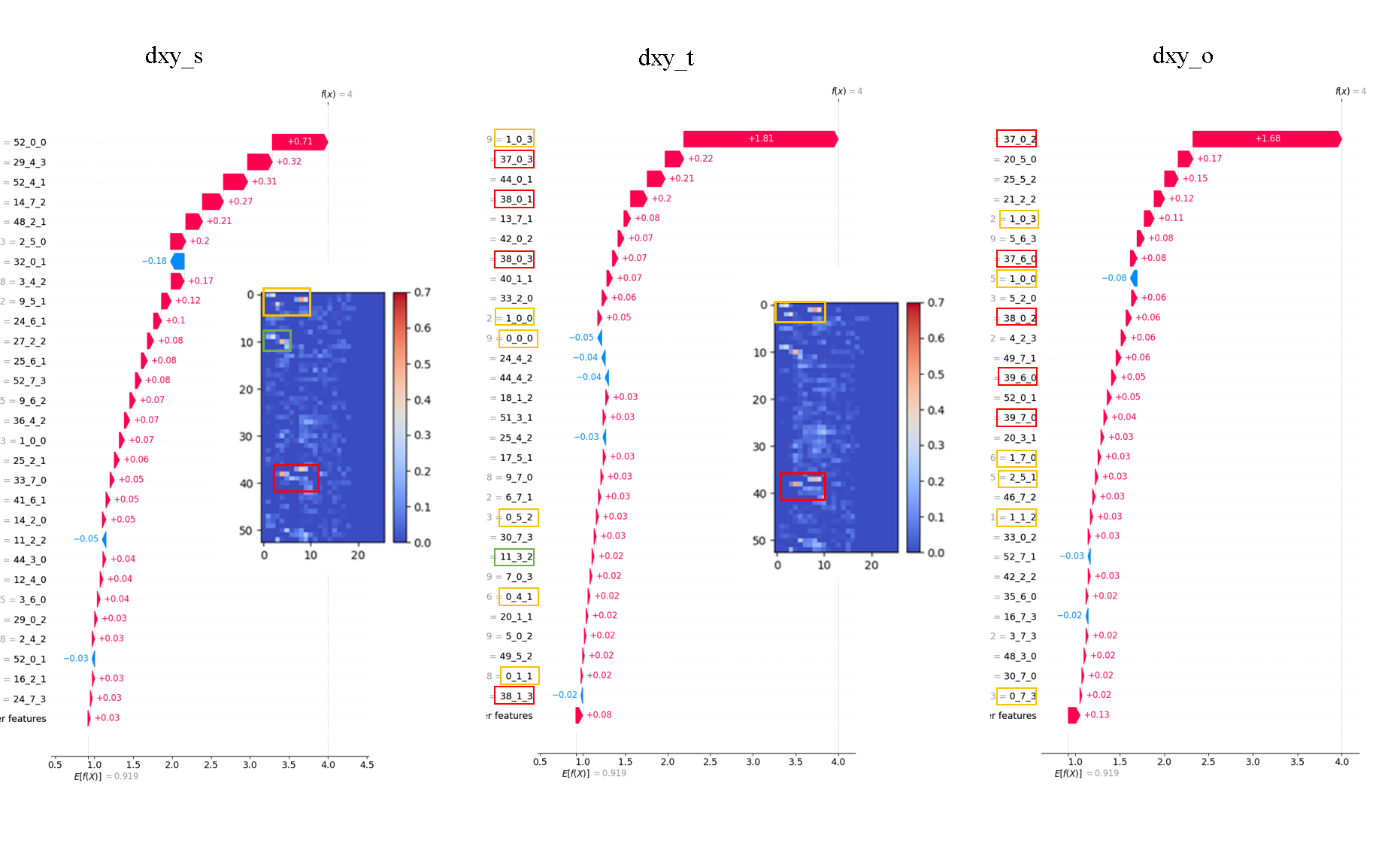}
    \caption{Waterpool picture of HbR. Y-axis denotes feature in format '\textbf{Brain channel}\_\textbf{Topk amplitude indices}\_\textbf{Feature dim}'. X-axis denotes SHAP values. Red represents forward gain and blue represents backward gain. We also use red, green, yellow block to sign significant brain channels during Point-Biserial Correlation Analyses and SHAP analyses to show their correspondence in order to verify interpretable of our approach.}
    \label{fig:HbR_SHAP}
\end{figure*}
\begin{figure*}[!t]
    \centering
    \includegraphics[width=1\linewidth]{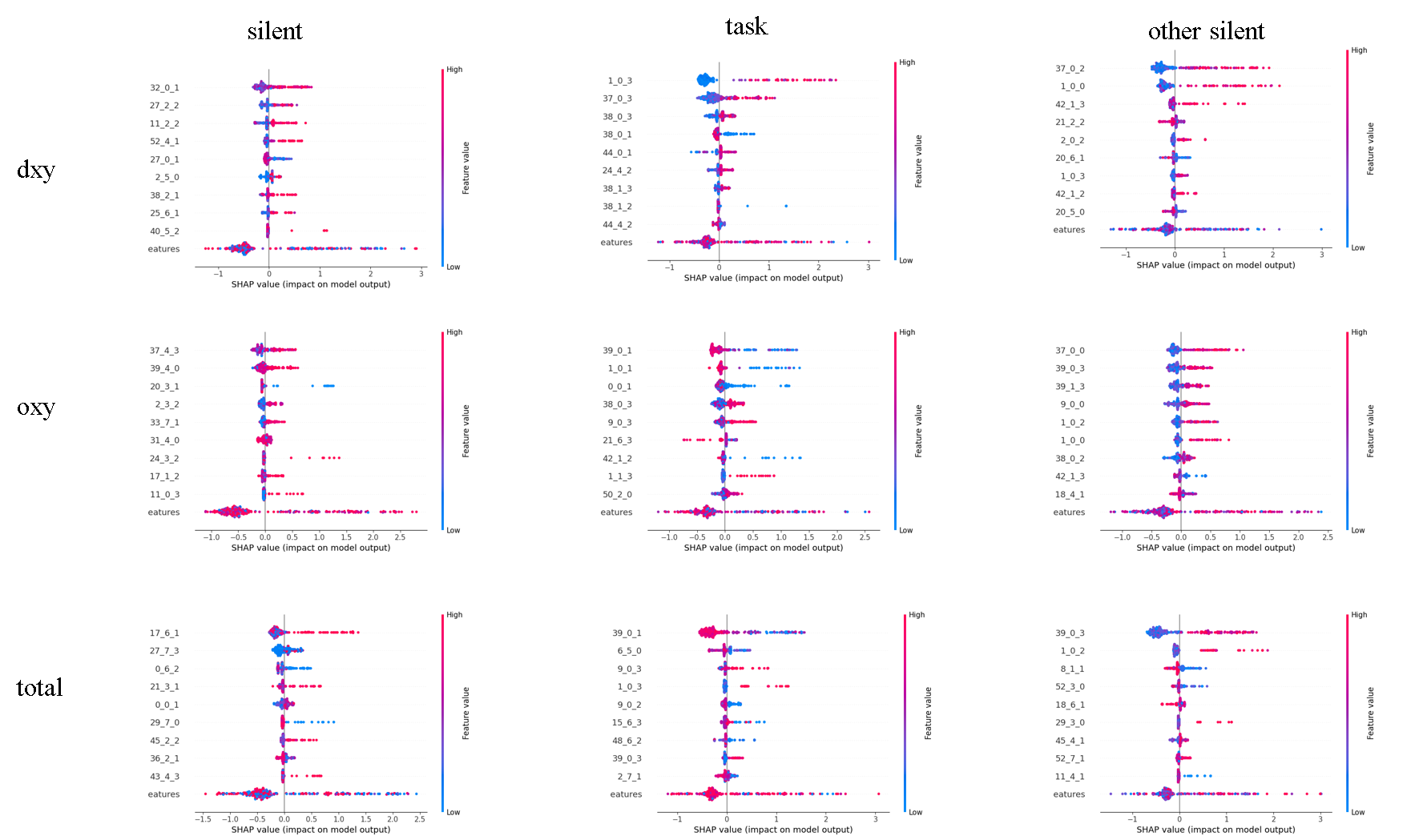}
    \caption{Beeswarm of top 30 biomarkers from global level. 0-2, 9-11, 37-40 brain channels show large contributions on depression dignosis, which fits previous medical research.}
    \label{fig:total explainer}
\end{figure*}
\subsection{Interpretability}
        Once we have trained our model, we would get the middle output of the network about our DFTs $f_{d}$. In clinical application, the task of depression detection requires that the model be interpretable. SHAP\cite{lundberg2017unified} is a relatively versatile method of model interpretability which constructs an additive explanatory model. It explains the "black box model" from both local and global levels by calculating the marginal contribution of features to the output of the model. Now, consider $f_{d}$ as $X$, classification labels as $Y$ and explore which $x_{i}$$\in$$X$ contributes to $Y$ via a XGBoost Explainer of SHAP.\par
        Fig \ref{fig:HbO_SHAP}  and  \ref{fig:HbR_SHAP} illustrate a wealth of pictures on the influence of the top 30 biomarker characteristics at the single sample level. We also plot a heatmap of the point-biseral correlation in Fig. \ref{fig:point-biseral} on the left side of the waterfall pictures. We frame the corresponding brain channel in both waterfall and point-biseral pictures with the same color.
        It is easy to notice that almost all brain regions corresponding to higher $|r|$ values in the left image appear in the features of the right figure, indicating that our model could have learned effective our new biomarker characteristic, as we expected. Moreover, we found that both HbO and HbR in the silent period have a lower absolute maximum value of SHAP than in task period and other silent period, which confirmed that compared to the silent period, the task and other silent periods contribute more to classification labels. Above all, it has proved it is efficient to model each period independently because modeling all periods as one may bring noise from the silent period into the task and other silent phases.\par
Fig. \ref{fig:total explainer} demonstrates beeswarm: a scatter plot for top 30 features' density from the global level. Firstly, we could observe that features of tasks and other silent periods are wider than those of silent periods, indicating that tasks and other silent periods contribute much more than the silent period. Moreover, we also found in Fig \ref{fig:total explainer} of columns 2 and 3 (corresponding to task and other silent periods), significant features are almost gathered in the brain channel 0-2, 9-11, 37-40. This finding is consistent with previous medical research\cite{wang2021depression}. We can also find what frequencies of brain channels play an important role in classification from Fig. \ref{fig:total explainer}.
\section{Conclusion}
        In this paper, we first made a new larger fNIRS depression diagnosis dataset, which is the largest to our knowledge. Based on the dataset, we proposed a new biomarker and a TGCN-based model tallor-made for it, which got the highest performance compared with other methods. Finally, we managed to explain the features based on our novel biomarker. In summary, our work has solved some limitations of previous work, both in terms of dataset and model performance.\par
        The limitation of our work is that we have not tried to use multimodal data (SCL-90\cite{wang2024unraveling}, EEG, video and linguistic) to classify yet and we have not tried a multicategorical task for the diagnosis of depression\cite{seo2023multi} yet. The above two points will become the direction of our future research.\par
\section{Reference}
\bibliographystyle{ieeetr}
\small\bibliography{reference}
\onecolumn
\end{document}